# A Conclusive Remark on Linguistic Theorizing and Language Modeling

Cristiano Chesi

NeTS-IUSS Pavia, Italy <cristiano.chesi@iusspavia.it>

Considering the proliferation of responses to Piantadosi's original paper and the ongoing debate sparked by this special issue of the Italian Journal of Linguistics, it is clear that the discussion has touched a raw nerve in linguistic theorizing.

In the original target paper (Chesi, this issue), I illustrated three prototypical (and in many respects, extreme) positions — the computational, theoretical, and experimental perspectives — without explicitly endorsing any of them. Instead, I attempted to highlight what I believe are the key weaknesses of each of these prototypical stances, ultimately concluding that formal (i.e., 'generative') linguistics — more specifically, Minimalism, my theoretical comfort zone — must adopt practices and tools that are common in both computational and experimental fields.

As noted by most respondents, the title and some of the more extreme statements were intended as mild provocations to draw attention to core issues affecting linguistic theorizing. My position — somehow obscured behind the 'three-body problem' — is that any relevant scientific progress is driven by theoretical insight, not by trawling using experimental or computational methods that are cost-inefficient, energy-intensive, and ultimately unsustainable. Moreover, in full agreement with most of the replies, I believe that the success of certain large language models (LLMs), which are based on specific architectural assumptions, does not constitute a refutation of the generative paradigm. On the contrary, it strongly supports several key intuitions that have emerged within the generative linguistic tradition (Rizzi this issue). However, a concrete problem of 'incommensurability' arises (Hao this issue), as differing methodologies and specialized jargon (Butt this issue) often result in circular, unresolved discussions.

Before turning to the core of my final remark, let me first clarify a widespread confusion found in many of the critiques of Piantadosi's position that 'LLMs express Theories'. A general attitude in the critical literature toward this statement reflects a criticism oriented towards LLM models, rather than a critique of model architecture. The difference is substantial, as no generative linguist would ever conflate an adult's actual grammar with the Language Acquisition Device (LAD) that enabled the individual to attain a mature state of linguistic competence. Exactly in the same way, LLMs express something very similar to adults'-like mature competence. I believe the parallel between the LAD and the network architecture used to train the language model is fundamental for understanding in what sense LLMs might indeed meaningfully express theories. From this perspective (Baroni 2022), network architectures can be seen as potential implementations of structural intuitions (or 'inductive biases', Goyal & Bengio, 2022) and deserve far more respect than they are typically afforded by prominent generative linguists, who reductively criticize LLMs without seriously addressing the architectural and training-regimen factors that enabled the development of these models. Fortunately, some of the more substantive critiques of the 'LLMs express theories' position are included in the replies (e.g., Onea et al., this issue), and we should take this opportunity to present well-grounded perspectives on the matter.

Several important points have been raised in the replies, which, in the spirit of constructive and productive dialogue, I will attempt to summarize in three arguments that may help readers navigate the range of perspectives. Those are:

1. Lack of Explanation (LoE) argument: LLMs generate predictions that are not grounded in intelligible or consistent theories or hypotheses, raising concerns about their explanatory value.

2. Lack of Formalization (LoF) justification: While intuitions must eventually be precisely formulated, preliminary ideas can still significantly advance research before being fully formalized and coherently integrated into a theoretical framework. Moreover, theoretical disagreement often signals vitality and the ongoing evolution of a paradigm.

3. Divergent Goals (Goal) argument: Generative linguistics is committed to a cognitive perspective, whereas LLMs are primarily commercial tools developed to address computational downstream tasks.

I will briefly address each of these points in the following sections.

## 1. The Lack of Explanation (LoE) Argument

Almost all replies concur that *simulation* is not equivalent to *explanation*. LLMs effectively mimic linguistic production, but (§1.1) although they may operate within the same computational domain (i.e., achieve observational adequacy), the level of description they provide is fundamentally disconnected from the one required to account for the human linguistic faculty (Ginsburg this issue). Moreover, the mechanisms by which LLMs generate language are largely unintelligible, which precludes them from being considered theories in the strict sense (Ginsburg this issue; Onea, Kobayashi, and Wurmbrand this issue; Rizzi this issue). In addition (§1.2), the architectures underlying LLMs often disregard basic grammatical intuitions, particularly structural-dependency constraints that are central to generative theory (Fong this issue; Ginsburg this issue; Ramchand this issue). These two points deserve further elaboration.

### 1.1. On Intelligibility: From the Computational Level to the Algorithmic Level

In the target paper, I made a careful effort to articulate all three levels of adequacy in a manner that is both measurable and comparable. While it is tautological that a model capable of generating any sort of grammatical sentences to which it has never been exposed must be considered *observationally adequate* with respect to language L, the absence of explicit 'rules' at certain levels does not entail the absence of an intelligible deductive apparatus at another level. For this reason, I decided to provide a definition of *Descriptive Adequacy* that simply measures the theory size and puts in background the notion of *intelligibility* (Rizzi this issue). While we all implicitly agree on the fact that a theory is useful and elegant if it is simple and understandable in each of its deductive steps, much debate has arisen from the intuition that this 'simplicity' and 'understandability' might not be formulable at certain high levels, but only at lower ones. In the history of science, it is clear how the scope of 'intelligibility' has been reduced — from Galileo's Scientific Revolution, to Chomsky's Cognitive Revolution (Chomsky 2012), and more recently, to contemporary causal accounts framed in terms of 'functional intelligibility' (Cao and Yamins 2024). Returning to the connectionist discussion, 'linguistic complexity' was often regarded as an emergent property, with the only simple and intelligible level being that of the lowest artificial neural network (ANN) units and connections (Rumelhart, McClelland, and PDP Research Group 1986) — the algorithmic level. According to the Minimalist framework, simplicity must be described at the level of abstract structure-building operations — the computational level. This difference in levels of analysis often leads to misunderstanding, as it involves the comparison of two fundamentally 'incommensurable' intuitions (Hao this issue) unless linking hypotheses are explicitly formulated. I am firmly convinced that all experiments conducted since the early works of Rumelhart, McClelland, Hinton, and colleagues — particularly those addressing core linguistic problems such as English past tense acquisition (Rumelhart and McClelland 1986) — have yielded meaningful results only when specific architectural solutions were adopted. For instance, the successful modeling of past tense formation required a specific phonetic (Wickelphones, Wickelgren, 1969) encoding of the input paired with a particular ANN architecture: a pattern associator network (McClelland and Rumelhart 1991). Despite certain naiveties in the early experiments — such as inconsistencies in training regimens and learning trends interpretation — subsequent research continued and began to explicitly address 'cognitive plausibility,' understood as the implementation of specific architectural constraints designed to enhance the performance of these models (Kirov and Cotterell 2018).[1]

In this spirit, the notion of '(genuine) explanation' — often invoked without providing measurable criteria — appears to be overused in recent generative papers (Chomsky et al.

---

[1] An alternative perspective bears on the notion of 'representational convergence' (Huh et al. 2024): despite differences in architecture, training regimens, and data modalities, models addressing complex problems tend to converge on remarkably similar representations of data points. While this line of inquiry likely requires more rigorous linguistic investigation — the 'color' experiment, for instance, is very limited — the hypotheses proposed to explain convergence are nonetheless compelling. These include the 'simplicity bias', 'multitask scaling', and 'capacity' hypotheses, which might explain why larger models are more likely to converge than smaller ones.

2023, XX). As Chomsky admitted (Chomsky 1968), this is a rather slippery concept and a further source of potential 'incommensurability' (Hao this issue).[2]

Intuitively, I agree with all the respondents who have highlighted that notions such as C-command possess explanatory power (Rizzi this issue). However, if we aim to measure this 'explanatory power', the only viable approach is to compare the *simplicity* (Descriptive Adequacy) and *efficiency* (Explanatory Adequacy) of a theoretical model by assessing how the implementation of such constraints functions within a fully developed theory. For instance, C-command can be formalized algorithmically as a derivational constraint that applies at each Merge operation, or alternatively, it can be implemented as a representational filter applied to a structural fragment before or after Spell-Out. I do not believe these two approaches will prove equivalent — neither in terms of theory size (*simplicity* or Descriptive Adequacy), nor in terms of effective acquisitional constraints (*efficiency* or Explanatory Adequacy).

With regard to the notion of 'theory size' it is worth noting that Chomsky began engaging with a related concept of simplicity in his early work — as briefly mentioned by (Haspelmath this issue) —, and more recently revisited the issue from a critical standpoint (Chomsky 2021). In this context, the example he offers is illuminating, though arguably in the opposite direction from what he intends: rather than supporting the claim that "size should not matter," it may highlight the relevance of theoretical economy.

Briefly, Chomsky (2021, p. 7) proposes two Context-Free Grammars, G1 and G2 as illustrated in (1). These grammars are identical in size, having the same number of rules and symbols. He argues, however, that only G1 provides a 'genuine explanation' by capturing the optionality of the element B (an adjunct, such as in the phrase: [$_Y$ read] [$_W$ the book] [$_B$ in the library])

(1)  G1 = {X → YWB, X → YW}

   G2 = {X → BWY, X → YW}

Indeed, under even a basic implementation of the Minimum Description Length (MDL) intuitions — using a simple size-based metric — we can already predict the superiority of G1 over G2. This is because G1 is *compressible*, as YW is a shared prefix of the rewritten part, whereas G2 lacks such redundancy. As a result, G1 can be encoded in a more compact form (G1 = {X → YW(B)}), aligning with the criterion of superior *descriptive adequacy* (Chesi 2025, 8). From this perspective, the MDL objective subsumes the incommensurable notion of 'genuine explanation' in a quantifiable manner.

---

[2] In his reply to Quine's work (Quine 1960), Chomsky explicitly criticized the reference to a 'genuine explanation' when it was invoked to support the internally inconsistent stance of behaviorism (Chomsky 1968, n. 11).

Graf (Graf this issue) is the only contributor in the replies who seriously challenges the assumptions underlying the MDL framework. He criticizes MDL metrics, particularly in cases where the 'corpus cost' — used as a proxy for the computational domain to approximate observational adequacy — makes the 'grammar cost' negligible (Ermolaeva 2023). This is a concrete risk that can be mitigated only by adopting a *rationalist*, rather than *empiricist*, approach — as previously argued (Chesi 2025, 31). We do not need "a bigger corpus" to demonstrate that a theory is observationally and descriptively more adequate; a few additional contrasts that selectively challenge specific theoretical assumptions are sufficient[3]. As Marantz (2019, p. 8) puts it, "puzzles are counterexamples to predictive linguistic theories that arise in the absence of an alternative theory that predicted them." All remaining data can, at least provisionally, be treated as 'data dust' — data that, while not immediately relevant, may gain significance in light of future theoretical developments (Wiltschko this issue). Continuing with another illuminating claim by Marantz (2019), as cited in Onea et al. (2025): "Linguists predict data they don't have, the body of empirical generalizations uncovered by the methodology grows year by year, and alternative accounts of phenomena are in fact pitted against each other, with THE LOSERS NO LONGER VIABLE [emphasis added]". I would fully endorse Marantz's claim, and I sincerely hope he is correct. However, I am compelled to observe that various blatantly incorrect generalizations continue to be propagated from one peer-reviewed paper to another, often without contributing any clear theoretical advancement (Butt this issue).

---

[3] Here again, Graf identifies a logical counterargument (see §2.1): infinitely many contrasts could, in principle, be included in this respect. This is true, but only if we overlook the fact that each relevant contrast must target a specific assumption that differentiates two competing theories. Infinitely many contrasts would be required only if there were infinitely many theories making distinct predictions about contrasts that cannot be dismissed as mere 'data dust'. However, such a scenario would make language acquisition 'in the limit' (Gold 1967) logically impossible.

Unfortunately, none of the respondents explicitly addresses the (speculative) notion of 'data dust.' Stabler engages with a related issue, rephrasing the problem of 'dust under the carpet' as a matter of 'dust on the lenses' (Stabler this issue, 7). However one chooses to frame it, certain confounding data that fail to align with any coherent hypothesis must be set aside. Temporarily disregarding such data allows linguists to concentrate on promising contrasts (Rizzi this issue), — unless, of course, those data are systematically organized and thereby transformed into linguistic puzzles, in the sense articulated by Marantz (2019).

Generally, it holds for any set of linguistic data that the most efficient theory — the one capable of predicting all grammatical sentences — will also be the one that most effectively and precisely compresses the reference corpus. However, we are not seeking a lossless, zip-like compression algorithm (Katz 1986). Our theory should not preserve irrelevant details such as the exact wordings and orderings of individual sentences. Instead, we require a lossy algorithm — one that discards a substantial amount of information while retaining the ability to generate and recognize only grammatical sentences. To be clear, our interest in LLMs should not stem from their ability to pass the Turing Test, but rather from the architectural and training assumptions that enable them to consistently generate grammatical sentences and make human-like judgments about grammaticality through minimal pair comparisons.

## 1.2. On Grammatical Intuitions

It is true that most architectures do not incorporate relevant linguistic considerations into their design (but see §1.1 and footnote 1). Notably, the widely adopted attention-based mechanism — fostered by the (possibly 'social,' in the sense of Hao, 2025) success of the unpeer-reviewed paper by Vaswani et al. (2017) — bears little resemblance to 'attention' as understood in psycholinguistics, nor does it reflect relevant linguistic structural constraints (Fong this issue; Onea, Kobayashi, and Wurmbrand this issue). Similarly, the training algorithms appear implausible in many respects (Ramchand this issue).[4]

However, this does not imply that grammatical principles cannot be integrated into a network architecture, thereby constraining the computational flow according to specific structural intuitions (Sartran et al. 2022). To clarify this point, consider modifying a specific gate within a computational graph to implement the operation Merge[5]. This may appear complex, but it is essentially what we do when fully formalizing a structure-building operation such as Merge. Consider the first Merge example discussed in the target paper, here repeated:

(1)     *Merge*(scolds, Bill) = {scolds, Bill}

---

[4] See Lillicrap et al. (2020) for a broader perspective.

Any ANN architecture simply encodes <scolds> and <Bill> (or the sub-tokens from which they are composed) as vectors — i.e., sequences of numbers representing abstract features relevant to the intended computation (Butt this issue; Onea, Kobayashi, and Wurmbrand this issue). These vectorial representations — commonly referred to as 'word embeddings' in computational jargon — may allocate specific components to encode a word's categorial status, relevant agreement features (e.g., person and number), and selectional properties. For instance, one component might represent external argument selection, another internal argument selection, with additional components potentially encoding further categorical or semantic specifications. In the end, 'Bill' and 'scolds' will be represented by vectors such as <1, 0, …, 1> or <1, 0, …, 0>, which can be approximately interpreted at an intelligible level as <V, N, …, person, number, … > as in (2)[6]:

(2)   scolds =   <1,   0, …, 0,   0,     1,      1,             … >
      Bill =     <0,   1, …, 0,   0,     0,      0,             … >
                 <V,   N, …, pers, numb, arg_external, arg_internal, … >

While I fully agree with the point made by Onea et al. (2025) — particularly with regard to the opaque manner in which 'vectorialization' is implemented as a necessary preliminary step for compressing lexical input and reducing the number of early parameters to be tuned[7] — the fundamental question a generative linguist might ask is how Merge might operate on these representations. For the purposes of this inquiry, we can reasonably assume that a meaningful vectorial representation is manually provided for each relevant lexical item, thereby bypassing the preliminary parameters associated with vectorialization or word embedding. This approach makes explicit the hypotheses about what should be represented within the 'black box' (Ginsburg this issue; Onea, Kobayashi, and Wurmbrand this issue; Zamparelli this issue) and what should instead be subject to learning.[8] For example, in the BabyLM 2024 Challenge, we evaluated various mathematical hypotheses — such as vector summation, concatenation followed by a sigmoid transformation, dot product, and point-wise multiplication — and ultimately found that concatenation combined with a sigmoid transformation[9] best implemented a specific version of a unification procedure, a well-known concept within the HPSG framework (Pollard and Sag 1994), yielding the most accurate predictions (Chesi et al. 2024). The results of our experiments barely reached the baseline on the BLiMP benchmark; however, we demonstrated that modifying the gating mechanism in this way induces some degree of linguistic coherence in the behavior of our Small Language Model, which was trained on a naturalistic, child-directed corpus. Specifically, the trained model consistently selected either the grammatical or the ungrammatical item over 80% of the time across groups of minimal pairs representing specific linguistic phenomena. It is important to highlight that an overall accuracy around 50% may reflect either entirely random performance — for instance, when the model achieves approximately 50% accuracy across groups of phenomena that differ only in irrelevant lexical variation within a given syntactic pattern — or a coherent linguistic behavior, such as 90% accuracy on half of the phenomena and only 10% on the other half (consistently selecting the ungrammatical alternative). Overall, both models perform at approximately 50% accuracy; however, only the second model exhibits a sufficient linguistic consistency. I believe that these kinds of 'technical solutions' offer effective linking hypotheses between the computational and algorithmic levels, as envisioned by many (Butt this issue; Ramchand this issue), potentially bridging the gap between linguistic theorizing and language modeling in a productive way — likely aligning with the expectations for small-scale experiments proposed by (Onea, Kobayashi, and Wurmbrand this issue). It is important to remember that, to conclude that our theory is explanatorily adequate, we ask more than simple implementations of structure-building operations: the model must also be able to bootstrap from Primary Linguistic Data (PLD) and consistently exhibit adult-like performance after reasonable exposure to such input. Without seriously engaging with the acquisitional perspective — in generative linguistic terms, or 'training' in computational terms — a theory can, at best, be considered *descriptively adequate*.

---

[5] A gate is a simple mathematical operation that combines one or more inputs into a single output.

[6] Once the network has been trained, various analytical techniques — such as Principal Component Analysis (Elman 1990) — can be used to investigate which types of features are encoded in the embeddings.

[7] I would also include the 'tokenization' step in this readily criticizable standard pipeline (Fusco et al. 2024).

---

[8] This is not the only possible approach. One might instead adopt a bottom-up perspective, in which LLMs are probed to infer their more-or-less categorical internal representations (Baroni 2022; Zamparelli this issue).

[9] I.e., we created a vector of size double than the original word embedding vector, concatenating the first word embedding with the second, then we squeezed this long vector to its original embedding size using a sigmoid transformation.

*On Unboundedness*

The lack of "grammatical intuitions" in LLMs has also been supported by a separate argument. According to several scholars (Collins 2024; Onea, Kobayashi, and Wurmbrand this issue; Ramchand this issue), LLMs — or more precisely, ANNs — do not qualify as linguistic theories, since they are universal function approximators (Hornik, Stinchcombe, and White 1989), that is, they can simulate Turing-equivalent computations. In my opinion, this is a risky argument: by the same logic, one could also criticize phrase structure grammar (Chomsky 1957), since unconstrained rewriting rules yield Turing-equivalent computational power. Much of Chomsky's early work focused on identifying the relevant constraints that limit the domain of computation to what is necessary for capturing core linguistic phenomena. There is now a broad consensus that mild context-sensitivity (Shieber 1985) is likely necessary to capture most relevant linguistic properties. If certain scholars are correct, this also applies to ANNs, with the perhaps unsurprising conclusion that Recurrent Neural Networks (RNNs) — due to their inherent recursive mechanisms — are better suited to capturing truly infinite recursion than transformers (Delétang et al. 2022). For years, Chomsky's hierarchy has served as a foundational tool for researchers introducing new grammatical frameworks, allowing them to demonstrate equivalence with existing formalisms within the mildly context-sensitive domain (Butt this issue; Graf this issue). The rationale was that, since we have efficient algorithms (with polynomial time complexity) for recognition and generation within this domain, we can be confident that our grammatical formalism is both computable and sufficiently rich and efficient. In fact, some scholars have attempted to show that, by applying different constraints to various ANN architectures, it is possible to obtain models with different computational powers that align with levels of Chomsky's hierarchy (Delétang et al. 2022) or express them in terms of circuits complexity (Merrill, Sabharwal, and Smith 2022). From this perspective, Baroni's argument (Baroni 2022) is entirely reasonable: ANNs define a space of possible grammars, and it might be a matter of architectural constraints to determine the relevant boundaries within that space.

## 2. *The Lack of Formalization (LoF) Justification*

As many have noted, the paper focused exclusively on Minimalism as the most recent iteration of the generative enterprise, and my critiques — specifically concerning 'personal Minimalism' — are confined to that framework. Other grammatical frameworks are arguably in a stronger position with respect to formalization, including LFG (Butt this issue), HPSG, and TAG, among others. Additionally, emerging approaches appear promising in this regard (Ginsburg and Fong 2019; Graf this issue; Stabler 2013), particularly given the ability to capture the 'gradual' dimension of relevant linguistic data (Butt this issue).

The vast majority of responses share the view of language as a 'computational system' (Fong this issue) that can be fully formalized. While this is a trivial consequence of adopting any explicit computational model, it does not imply that every linguistic intuition can — or should — be formally specified from the outset.

I agree with Hao in this respect: "formal vagueness and data idealization can enhance a reader's feeling of understanding, and in turn, increase a theory's perceived explanatory power." (Hao this issue, 12). I am also well aware of the intricacies that led (Graf this issue; Haspelmath this issue; Stabler this issue) to critically examine and reformulate some of my positions in order to articulate a more productive and attainable challenge for linguistic theorizing. Without going into the details of their alternative proposals, I would like to address just two fundamental points: The first is the risk of subordinating theory to data — a concern captured by the 'dataism' (Ramchand this issue) and 'benchmarkification' issues (Graf this issue), §2.1. The second point is related to the lack of technical expertise needed to adequately address the problem of formalization, §2.2.

### 2.1. *Dataism and Benchmarkification*

In response to a question about Popper's influence (Popper 1934) on his work[10] Chomsky emphasized the importance of determining what constitutes relevant data for linguistic theorizing: A data fragment that elucidates our cognitive faculty is not something that can be obtained through random or extensive data collection, but rather through a direct challenge of a theoretical assumption. On the same page, Rizzi emphasizes the fundamental right to exclude certain data points from the observational domain (Rizzi this issue) precisely with the intent of removing what Stabler calls the "dust under the lens" (Stabler this issue). This is the "dataism" threat, in other terms (Ramchand this issue). I fully endorse this perspective, which I consider both legitimate and necessary. It is, however, important to acknowledge that if a theory fragment $F_1$, which accounts for a verified generalization $G_1$, is inconsistent with a another theory fragment $F_2$, which accounts for a separate verified generalization $G_2$ — and neither $F_1$ nor $F_2$ can derive the generalization covered by the other fragment — then both should be set aside in favor of another theory fragment $F_x$ that accounts for both $G_1$ and $G_2$. Such a fragment should be considered more adequate, as it eliminates part of the 'dust under the lens,' or, from a more critical and possibly radical perspective, uncovers the 'dust under the carpet.' Clarifying this basic point was the purpose of my (possibly unfortunate) example in the target paper, where I discussed modularity by illustrating the sensitivity of extraposition to quantifier raising (Guéron and May 1984). Specifically, if a theory fragment $F_1$ (e.g., optional quantifier raising) is assumed to operate within one encapsulated module external to core syntax (e.g., LF), but must also affect the output of another distinct encapsulated module external to core syntax (e.g., PF), as predicted by theory fragment $F_2$, then $F_1$ is inconsistent with $F_2$ — assuming a strictly modular T-model architecture, in which the only point of contact between the two external modules is mediated through core syntax.

---

[10] https://www.youtube.com/watch?v=-xerglwYdkE

Importantly, while there is little doubt that 'eliminative reductionism' is an unproductive approach (Rizzi this issue), it remains crucial to recognize that a more constructive form of 'reductionism' (e.g., replacing $F_1$ and $F_2$ with a suitable $F_x$) is essential for eliminating unnecessary or inconsistent theoretical machinery, as sponsored by the notion of *perfection* in Minimalist terms. This includes the removal of mechanisms that overgenerate and require constraints imposed by multiple theory fragments, which are not mutually consistent. In this sense, I am increasingly convinced that a shared set of critical contrasts — on which linguists might broadly agree — constitutes our "Hilbert's list for syntax" or, even better, the empirical foundation that any theory aiming at both *descriptive* and *explanatory adequacy* must address. We might call this a 'benchmark,' or simply refer to it as textbook linguistic examples. As noted in many responses, and as emphasized in the target paper, all items included in widely used benchmarks such as SyntaxGym and BLiMP are drawn from generative linguistics publications. It is, frankly, somewhat disconcerting to encounter so-called 'theoretical solutions' that overlook empirical evidence discussed as early as the 1970s and 1980s.

As Graf (this issue) observes, 'benchmarkification' may have serious unintended consequences. On one hand, it can foster the so-called Matthew Effect, whereby more popular theories receive increasing support, monopolizing attention and resources at the expense of less widely endorsed alternatives. On the other hand, it may lead to a narrow focus on accuracy, as seen in parsing evaluation, where systems that perform well on high-frequency structures are favored — while low-frequency phenomena, which often serve as the core test bed for theoretical insights, are neglected (Fong this issue).

However, not all challenges are alike, and we risk missing a valuable opportunity for productive engagement if we exclude ourselves from novel initiatives such as the BabyLM Challenge[11]. It is true that adapting a model to perform on specific benchmarks can be very time-consuming. It is also true that many issues raised by current benchmarks may not align with the priorities of different research programs (Graf this issue). However, without shared priorities and a common ground for comparable predictions, 'personal Minimalism' risks perpetuating the fragmentation of research efforts. This fragmentation hinders the formation of a critical mass of researchers who could otherwise pursue alternative hypotheses within a shared evaluative framework. Ultimately, this also means missing the broader 'social challenge' and confining the field to clever but 'incommensurable' opinions (Hao this issue).

Small research communities must have the opportunity to demonstrate the value of their intuitions on a FAIR (Findability, Accessibility, Interoperability, and Reuse) ground (Wilkinson et al. 2016). I fully endorse Zamparelli's position in this regard (Zamparelli this issue): an open-source approach, designed to promote shared responsibility, will be vital for the future of generative linguistics and will help mitigate the risk of parochialism and theoretical incommensurability, possibly addressing both standardization (Zamparelli this issue) and categorical generalizations within a broader typological scope (Butt this issue; Haspelmath this issue; Wiltschko this issue). Without such an effort — essential for updating the "early simplistic version of UG" (Ramchand this issue) —, Haspelmath's concern (Haspelmath this issue) that, within the generative field, speculative ideas get promoted over scientifically grounded ones is likely to remain a pervasive perception.

---

[11] In the last 2024 Challenge, among the 29 paper accepted for the proceedings, just two mentioned some generative approach (Hu et al. 2024).

## 2.2. *Lack of Technical Skills and the Necessity of Mutual Support*

Many argued that generative linguists often lack the technical skills necessary to implement their models (Ginsburg this issue). At the same time, there is growing recognition of the need for mutual collaboration across disciplines (Ramchand this issue; Rizzi this issue; Stabler this issue; Zamparelli this issue). Since linguistic theories lie at the core of any meaningful benchmark designed to evaluate a model's or theory's coverage — or 'performance,' in computational terms — interdisciplinary teams will be essential. Teams that crucially include scholars capable of translating promising linguistic intuitions into algorithmic proposals that can be integrated into sound theoretical frameworks.

Implementing specific linguistic intuitions — priors, or inductive biases (Goyal and Bengio 2022) in machine learning terms— can offer valuable support for testing competing theories. Despite differing opinions on various technical aspects, we share a robust methodological foundation for addressing standardization challenges (Zamparelli this issue), which continues to ensure that the generative enterprise remains intellectually engaging under any well-defined formulation (Butt this issue). It is worth noting, however, that many model architectures and training datasets are now openly accessible (see, for example, the Hugging Face platform). Once again, there is little justification for ignoring how a linguistic intuition might be integrated into these openly available computational models through effective interdisciplinary collaborations.

## 3. *Divergent Goals: Embracing the Cognitive Stance*

An impressive comparison of learning efficiency in machine learning versus human learning is presented in (Fong this issue). This comparison highlights that our modest 20watt brain outperforms megawatt-consuming computational clusters in learning from limited data sets. It serves as a reminder that, while the goal of generative linguistics is to model how the linguistic faculty represents and processes linguistic input, LLMs are designed to perform downstream tasks such as machine translation (MT) or natural language understanding (NLU). Computational linguists developing LLMs often pursue the legitimate goal of improving performance on specific benchmarks for specific tasks (Papineni et al. 2001), frequently disregarding basic linguistic facts. I agree that, from the perspective of formal linguists, this can be seen as a frustrating and reductive objective, offering little or no return in terms of linguistic theorizing (Fong this issue; Ginsburg this issue).

It is also widely agreed that generative linguists adopt a 'cognitive stance' (Onea, Kobayashi, and Wurmbrand this issue; Rizzi this issue; Wiltschko this issue). It is, however, in the interest of linguists to demonstrate that, for certain tasks, incorporating specific linguistic intuitions can offer significant advantages — for example, in terms of efficiency, where low-resource models can be built with negligible drops in performance. Explanation — understood in the sense discussed in §1 — remains the central goal of generative linguistics. However, it is difficult not to be drawn to alternative objectives, such as identifying optimal architectures for processing specific linguistic properties (Lan et al. 2022). The notion of a 'perfect system' that has guided minimalist hypotheses (Chomsky et al. 2023, 55) must also contend with so-called 'third factors' — those related to computational efficiency — and invites further exploration of concrete cases involving interface conditions (Butt this issue).[12]

From the perspective of mutual support, one key role of generative linguistics is to identify confounds in test sets. For instance Kodner et al. (2023, p. 8) noted that BLiMP (Warstadt et al. 2020) contains significant shortcomings that may allow models to perform well on the benchmark without engaging in the kind of structural inference expected of humans (Graf this issue; Onea, Kobayashi, and Wurmbrand this issue). This is true; however, on the one hand, we can construct better minimal pairs by relying on more complex grammars instead of linear patterns (Bressan et al. 2025; Lan, Chemla, and Katzir 2024). On the other hand, we expect that inferring dependencies from linear patterns requires more parameters than doing so from hierarchical ones. In this case, the definition of 'Descriptive Adequacy' (Chesi 2025, 8) serves as an important safeguard, as it favors the most parameter-economical theory.

---

[12] A curious paradox that might arise, for instance, concerns 'linearization': even if we could hypothetically use telepathy to communicate (Chomsky 1995, 221), this would not significantly accelerate our exchanges nor eliminate the need to chunk linguistic information. This limitation stems from the *incrementality bottleneck*, which arises because the same portion of our finite 'linguistic organ' — however it is structured — must be reused to process incoming linguistic input that each time saturates our maximal capacity (Chesi forthcoming).

## 4. A Reconciliatory Perspective

Like many respondents (Stabler this issue), I am inclined to view disagreement and argumentation as signs of vitality within the field. Apparent inconsistencies may signal forthcoming paradigmatic shifts (Wiltschko this issue), suggesting that numerous promising directions are likely to emerge in the near future. If experimental, computational, and formal linguists join forces — under a unificational and open-source rather than eliminative reductionist perspective — many disagreements may be reframed as dialectical tensions, while other critical issues may be more productively addressed. We should remember that if "the singular purpose of generative linguistics [remains] to explain language to generativists" (Hao this issue, 8), then the widespread skepticism expressed by neighboring approaches will ultimately be justified (Haspelmath this issue).

Since non-trivial structural priors are embedded in any machine learning architecture (Baroni 2022; Goyal and Bengio 2022; Hao this issue), we can begin to meaningfully compare architectural assumptions and pose more substantive questions from a linguistic perspective. For instance, are there architectural specificities that specifically favor language acquisition over other cognitive functions? Is there a single algorithm that is demonstrably more efficient than others under certain learning circumstances? Returning to Chomsky's opening lecture at IUSS in 2012, by endorsing a computational or experimental perspective we need not revert to the 'mechanical philosophy' or 'empiricism' but simply adjust our expectations of 'intelligibility' in favor of greater theoretical consistency in terms of both measurable descriptive and explanatory adequacy.

I remain convinced that generative linguists will continue to develop theories inductively, deriving them from a relatively small set of observations and taking fruitful advantage of small-scale computational simulations — ultimately aiming to describe core principles within more consistent and solid frameworks. This is NOT the end of generative linguistics, but rather a timely and constructive theoretical reorientation.